\documentclass{article}

\usepackage[preprint]{neurips_2025}

\usepackage[utf8]{inputenc} 
\usepackage[T1]{fontenc}    
\usepackage{hyperref}       
\usepackage{url}            
\usepackage{booktabs}       
\usepackage{amsfonts}       
\usepackage{nicefrac}       
\usepackage{microtype}      
\usepackage{xcolor}         

\usepackage{graphicx}
\usepackage{natbib}
\usepackage{arydshln}

\usepackage{amsmath}
\usepackage{cleveref}
\usepackage{enumitem}

\definecolor{darkgreen}{rgb}{0.0, 0.5, 0.0}

\title{Projected Compression: Trainable Projection for Efficient Transformer Compression}

\author{%
Maciej Stefaniak$^{1}$\thanks{Core execution.} \quad Michał Krutul$^{1, 2}$ \quad Jan Małaśnicki$^{1, 2}$ \\
\quad \textbf{Maciej Pióro}$^{1, 3}$\ \quad \textbf{Jakub Krajewski}$^{1, 2, }$  \quad  \textbf{Sebastian Jaszczur}$^{1, 2}$ \\
\textbf{Marek Cygan}$^{1,4}$  \quad \textbf{Kamil Adamczewski}$^5$ \quad  \quad \textbf{Jan Ludziejewski}$^{1, 2, \thanks{Initial idea.}} $\\
$^1$University of Warsaw \quad $^2$ IDEAS NCBR \quad $^3$Polish Academy of Sciences \quad \\
$^4$Nomagic \quad $^5$Wroclaw University of Science and Technology  \\
\texttt{m.stefaniak8@uw.edu.pl}\\
}

\begin{document}

\maketitle

\begin{abstract}

Large language models have steadily increased in size to achieve improved performance; however, this growth has also led to greater inference time and computational demands. Consequently, there is rising interest in model size reduction methods. To address this issue, we propose \textbf{Projected Compression}, a novel model compression technique, that reduces model weights by utilizing projection modules. Specifically, we first train additional trainable projections weights and preserve access to all the original model parameters. Subsequently, these projections are merged into a lower-dimensional product matrix, resulting in a reduced-size standard Transformer-based model. Unlike alternative approaches that require additional computational overhead, our method matches the base model's per-token computation step in FLOPs. Experimental results show that Projected Compression outperforms the comparable hard pruning and retraining approach on higher quality models. Moreover, the performance margin scales well with the number of tokens.

\end{abstract}

\section{Introduction}

As large language models (LLMs), exceptionally good at Natural Language Processing (NLP) tasks \citep{mann2020language, raffel2020exploring, bommasani2021opportunities}, continue to be developed at increasing scale, their computational and memory requirements present growing challenges for deployment, experimentation, and fine-tuning ~\citep{cottier2024rising, singh2025survey}. Reducing the size and operational costs of the model while preserving quality is, therefore, essential to make LLMs more widely accessible to the research community and the public. Model compression techniques have emerged as a popular solution to this problem, with pruning remaining one of the most widely adopted approaches due to its simplicity and effectiveness. However, standard hard pruning methods suffer from an inherent limitation: once parameters are removed, their representational capacity is permanently lost, often leading to performance degradation. As a result, hard pruned models usually require additional retraining. Leading in compression efficiency are soft and dynamic pruning methods, which incorporate training and compression in the same step. 

\begin{figure}[!ht]
    \centering
    \includegraphics[width=\textwidth]{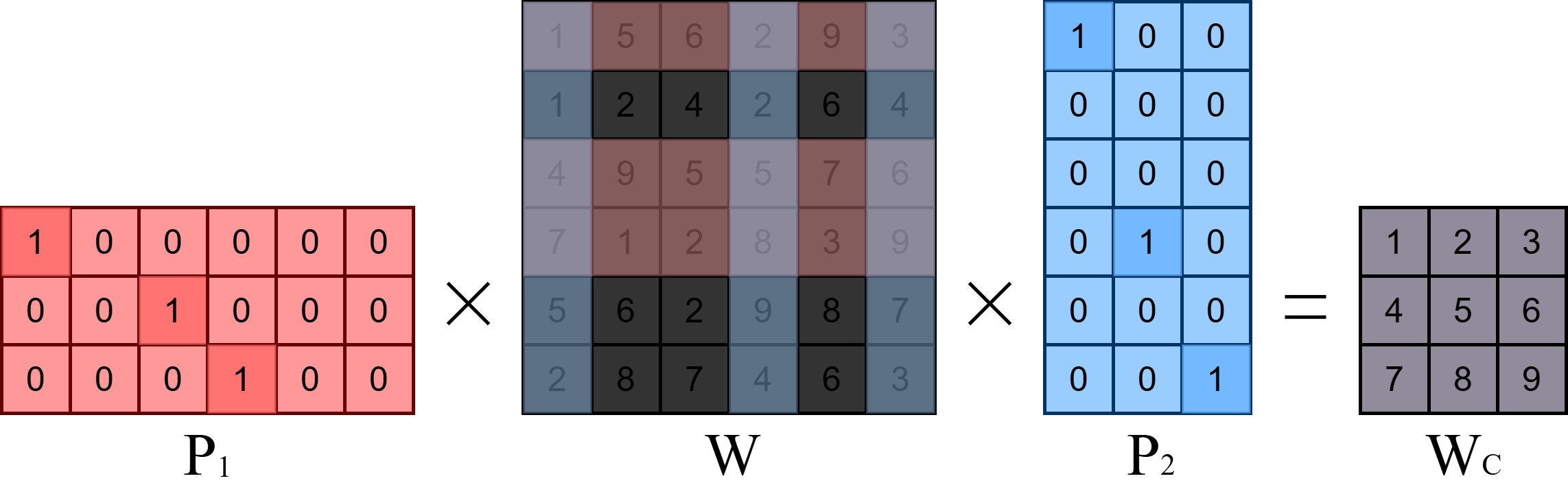}
    \caption{Simplified illustration of a projection module, where $P_1$ and $P_2$ are projection matrices, $W$ is a frozen base model parameters and $W_C$ is compressed weights matrix.}
    \label{fig:pc}
\end{figure}

Motivated by these advances, we propose Projected Compression (PC), a novel low-rank compression method that preserves access to all original model parameters through gradient-optimized projection modules. Rather than removing unimportant weights, PC redirects their influence through a learnable projection modules, which are initialized according to the importance of the base matrix weights. This enables the model to gradually reincorporate useful dimensions during training. This method retains the frozen base weights during training and allows compression by projection of their information, to the lower final dimensions of the compressed model - mixing their influence. We apply PC to Transformer-based language models and evaluate it under standard structured pruning settings. The preliminary results show that the PC improves performance against hard pruning and retraining given the same computational budget.

To our knowledge, PC is the first method to reframe model compression as a trainable projection problem operating on a frozen base model that takes advantage of additional weights importance information, explicitly targeting transformer core matrices under structured sparsity constraints. It offers a new cost-free alternative in hard pruning and retraining pipelines, decoupling the parameter elimination decision from the actual loss of information.

The main contributions of this work are:
\begin{itemize}
    \item Introduction of PC, a novel compression method that retains access to frozen base weights via trainable projection modules, while maintaining the same training cost per token as standard transformer. 
    \item Experimental results demonstraing performance gains of Projected Compression over hard pruning increase with: (1) quality of the base model relative to number of parameters, (2) base model size, and (3) training cost to obtain a compressed model.
\end{itemize}

\section{Related Work}

\paragraph*{\textbf{Hard pruning.}} A widely adopted technique for compressing deep neural networks, model pruning removes weights considered less important for model performance \citep{lecun1989optimal, hassibi1992second, han2015learning}. In transformer-based architectures, structured pruning that removes entire channels, attention heads, or layers has gained prominence due to its ability to produce actual inference speed-ups on modern hardware \citep{gale2019state}. Typically, pruning is followed by a retraining phase to recover lost accuracy. Hard pruning methods are based on importance criteria, most commonly magnitude-based thresholds \citep{han2015learning} or activation statistics \citep{hu2016network}. Although efficient and simple, these methods permanently discard information, limiting their ability to support aggressive sparsity without severe performance degradation. Projected Compression uses for initialization the same weight-importance criteria as hard pruning for parameters removal. 

\textbf{Dynamic and soft pruning} approaches mitigate sudden rigidity of information loss by allowing previously pruned connections to recover during training \citep{zhu2017prune, mocanu2018sparse, evci2020rigging}. In soft pruning, the masked weights remain in memory and continue to receive gradients, allowing their reemergence if they regain significance. Dynamic sparse training maintains a fixed number of active connections, reallocating them during training. These approaches outperform static hard pruning, but require specialized sparsity-aware training schedules and may suffer from overhead to irregular parameter usage during compression. More recent methods learn pruning patterns directly through differentiable masking or importance score learning \citep{xia2022structured, louizos2017learning, wen2016learning}. Although these approaches offer greater flexibility, they often require complex optimization. Projected Compression similarly enables all weights to influence model compression process mixing their influence by gradient optimized projections. 

\paragraph*{\textbf{Parameter-efficient fine-tuning.}} Parameter-efficient fine-tuning (PEFT) approaches, such as LoRA, Prefix Tuning, BitFit, and QLoRA \citep{hu2021lora, li2021prefix, ben2021bitfit, dettmers2023qlora}, enable model adaptation by freezing the base model weights and introducing a small set of trainable parameters. These methods are designed to efficiently adapt large models to downstream tasks under strict memory or computational constraints, often using lightweight modules such as low-rank adapters.
Projected Compression is conceptually related to LoRA and other PEFT approaches by freezing base weights and introducing trainable modules, projection matrices, which adapt the behavior of the projected model without modifying its base model core parameters.
It operates as a structured compression strategy aligned with the target for parameter reduction. PC does not aim to enable adaptability, but instead to create low-dimensional representations of the full model while maintaining access to its original capacity during training.

\paragraph*{\textbf{Other work.}}  In a concurrent work, Hao et al. \citep{hao2025tokenworth1000tokens} introduced a Low-Rank Clone (LRC) method for compression using a projection mechanism aligned with distillation during retraining. This operation propagates the full batch gradient through all projection matrix weights, making each step very costly. PC uses weights importance projections initialization similar to dimension reduction operation, which makes our method significantly cheaper.

\section{Method}
\subsection{Overview}

Projected Compression is a structured model compression method that uses \textit{projection modules} to construct a lower-dimensional representation of a Transformer model linear layers while retaining access to its full set of frozen base parameters. PC introduces additional learnable projection matrices that operate over the original weights. The compressed model is fully defined by these projections, which are gradient optimized during training, while the original base model weights remain frozen.

\subsection{Projection-Based Structured Compression}

Projected Compression targets structured dimension reductions in the Transformer architecture, specifically the \textit{feedforward hidden size} and \textit{embedding dimension} (model width). For each frozen base weight matrix $W \in \mathbb{R}^{d^{\text{in}} \times d^{\text{out}}}$, we are projecting $d^{\text{in}}$ and $d^{\text{out}}$ dimensions to a significantly smaller $d_S^{\text{in}}$ and $d_S^{\text{out}}$. Corresponding projection module is presented in \Cref{fig:pc}. This module consists of one or two trainable projection matrices, $P_1$ and/or $P_2$, which produce a compressed weight $W_C \in \mathbb{R}^{d_S^{\text{in}} \times d_S^{\text{out}}}$ of reduced rank:

\[
W_C = P_1 W P_2
\]
where:
\begin{itemize}
    \item $W \in \mathbb{R}^{d^{\text{in}} \times d^{\text{out}}}$ — the frozen base model weights,
    \item $P_1 \in \mathbb{R}^{d_S^{\text{in}} \times d^{\text{in}}}$ — a projection that downsamples the input dimension,
    \item $P_2 \in \mathbb{R}^{d^{\text{out}} \times d_S^{\text{out}}}$ — a projection that downsamples the output dimension,
    \item $W_C \in \mathbb{R}^{d_S^{\text{in}} \times d_S^{\text{out}}}$ — the resulting compressed weights.
\end{itemize}

In the one-sided variant, only $P_1$ or $P_2$ is used, depending on the compression axis.
During the course of training, only the projection matrices $P_1$ and $P_2$ are updated by gradient descent, while the base weights $W$ remains frozen. This allows frozen weights to be blended into the active computation subspace, offering the potential to recover and leverage information from parameters that would have been permanently removed in standard hard pruning. 

This mechanism preserves compatibility with the standard Transformer architecture. The projected weights $W_C$ are recomputed before each forward pass and are treated as regular layer parameters, ensuring full compatibility with the existing model infrastructure and methods such as an addition of distillation logits loss.

Importantly, this construction is functionally equivalent to a two-step transformation of the input token representation. For a given input token vector $x$, multiplying it by compressed weights $W_C$ (that is, computing $x W_C$) is equivalent to first projecting $x$ up into the base model space via $P_1$, performing a matrix multiplication with the full frozen weight matrix $W$, and then projecting the result back down via $P_2$:

\[
x W_C = ((x P_1) W) P_2
\]

This interpretation highlights that Projected Compression allows each token representation to interact with the full capacity of the base model during training, even though only a compressed projection is used during inference. It reinforces the intuition that PC does not discard any information a priori but instead passes token activations through learnable subspaces of the frozen parameter space.

\paragraph{Residual Weights Extension.} To improve late-stage flexibility during training, a residual term $W_r \in \mathbb{R}^{d_S^{\text{out}} \times d_S^{\text{in}}}$ is added to the compressed weights:

\[
W_C = P_1 W P_2 + W_r
\]

The residual weights $W_r$ are initialized with zeros and trained alongside $P_1$ and $P_2$. This addition helps the model maintain the optimization flexibility as it diverges from the frozen base.

\subsection{Initialization}

To align the initial compressed model with the base model, projection modules are initialized using importance scores derived from standard hard pruning heuristics (e.g., magnitude-based criteria). This ensures that the most important weights dominate the initial projection, while the less important dimensions are initially suppressed. Unlike hard pruning, however, Projected Compression does not eliminate any parameters - less important dimensions are excluded from the initial projection but remain accessible. As training progresses, these suppressed components gradually contribute to the model through optimization of the projection weights.

\subsection{Training Dynamics and Resource Efficiency}

One of the key advantages of Projected Compression is its training computing efficiency. When using a sufficiently large batch size commonly found in large language model training, the additional computational overhead introduced by projection modules can be omitted.

This efficiency comes from the behavior of modern automatic differentiation frameworks such as PyTorch \citep{paszke2019pytorch, paszke2017automatic}. During retraining, forward computation occurs only through the projected/compressed matrix. Gradient information is accumulated across batch and sequence dimensions in that smaller weight space. It makes it significantly cheaper during the backward pass, when only this reduced gradient information is used to update the projection weights $P$. The additional computation costs of managing projection modules become negligible for large batch sizes.

Consequently, the overall cost per optimization step is equivalent to that of retraining a transformer model obtained through hard pruning. However, this method comes at the expense of additional memory usage: the optimizer must store both the trainable projection matrices and the full set of frozen base model weights. This overhead can be mitigated through memory saving techniques such as gradient checkpointing \citep{chen2016training} and offloading inactive projection modules to CPU memory during training.

\section{Experiments}

\subsection{Experimental setup}
\paragraph*{\textbf{Pre-trained models.}} As an initial step, we trained a series of GPT-2 \cite{noauthororeditor} style  models to serve as base models for subsequent pruning experiments. Specifically, we used two model configurations: a 300M parameter model (16 layers, 16 attention heads) and an 800M parameter model (24 layers, 24 attention heads). For each configuration, models were trained on varying amounts of data to achieve token-to-parameter ratios of 20:1 and 80:1, respectively.

\paragraph*{\textbf{Evaluation.}} To assess the effectiveness of the proposed PC method, we conducted a comparative evaluation against the Hard Pruning baseline. For each of the pre-trained base models described above, covering two model sizes and two token-to-parameter ratios, we ran a series of experiments across varying numbers of training tokens and compression levels. Both pruning techniques were applied to each setting and performance was measured using the Cross-Entropy loss averaged over the last 100 training steps.

\paragraph*{\textbf{Training setting.}} Experiments were conducted across multiple clusters, with compute nodes equipped with 4× NVIDIA GH200, H100, or A100 GPUs. All models were trained on the C4 dataset \citep{raffel2020exploring} with the same sequences. 

\subsection{Results}
The experiments investigate scaling trends by comparing two compression pipelines: Hard Pruning with Retraining (HPR) and Projected Compression (PC). Both methods use identical weight-importance scores to guide pruning decisions and using the same number of training steps are matched in terms of compute.

\Cref{fig:pc_vs_hp} presents the main results comparing our proposed PC method with the Hard Pruning baseline at 50\% compression. Across all evaluated configurations, PC consistently yields lower Cross-Entropy loss than Hard Pruning on models trained with a higher token-to-parameter ratio and trained on more tokens. This trend highlights the strength of PC in preserving model quality under data-rich conditions. Additional results at other compression levels are provided in \Cref{tab:80tpp}, which further confirm the robustness of PC across a range of compression settings. All conducted experiments can be seen in \Cref{appx:a}.

\begin{table}[ht]
    \centering
\begin{tabular}{ccccccc}
\toprule  
& &  \multicolumn{5}{c}{Tokens processed during training compression pipeline}\\
Model $N$ & Method & \texttt{1.25B} & \texttt{2.5B} & \texttt{5B}& \texttt{7.5B} & \texttt{10B}\\
\bottomrule
\multicolumn{7}{c}{35\% compression}\\
\midrule
\text{300M} & \text{HPR} & $\textbf{3.1759}$ & $3.1362$ & $3.1070$ & $3.0925$ & $3.0775$\\
\text{300M} & \text{PC} & $3.1776$ & $\textbf{3.1361}$ & $\textbf{3.1037}$ & $\textbf{3.0882}$ & $\textbf{3.0711}$\\
\hdashline
\text{800M} & \text{HPR} & $\textbf{2.9224}$ & $2.8827$ & $2.8519$ & $2.8420$ & $2.8259$\\
\text{800M} & \text{PC} & $3.0327$ & $\textbf{2.8780}$ & $\textbf{2.8465}$ & $\textbf{2.8355}$ & $\textbf{2.8192}$\\

\toprule 
\multicolumn{7}{c}{50\% compression}\\
\midrule
\text{300M} & \text{HPR} & $\textbf{3.2688}$ & $\textbf{3.2172}$ & $3.1781$ & $3.1577$ & $3.1402$\\
\text{300M} & \text{PC} & $3.2823$ & $3.2198$ & $\textbf{3.1744}$ & $\textbf{3.1528}$ & $\textbf{3.1341}$\\
\hdashline
\text{800M} & \text{HPR} & $3.0286$ & $2.9733$ & $2.9300$ & $2.9128$ & $2.8930$\\
\text{800M} & \text{PC} & $\textbf{3,0217}$ & $\textbf{2.9652}$ & $\textbf{2.9203}$ & $\textbf{2.9037}$ & $\textbf{2.8840}$\\

\toprule 
\multicolumn{7}{c}{65\% compression}\\
\midrule
\text{300M} & \text{HPR} & $\textbf{3.3753}$ & $\textbf{3.3095}$ & $3.2603$ & $3.2392$ & $3.2175$\\
\text{300M} & \text{PC} & $3.3858$ & $3.3134$ & $\textbf{3.2591}$ & $\textbf{3.2341}$ & $\textbf{3.2133}$\\
\hdashline
\text{800M} & \text{HPR} & $\textbf{3.1245}$ & $\textbf{3.0569}$ & $\textbf{3.0056}$ & $2.9828$ & $2.9622$\\
\text{800M} & \text{PC} & $3.1375$ & $3.0626$ & $3.0069$ & $\textbf{2.9815}$ & $\textbf{2.9591}$\\

\bottomrule
\end{tabular}
\vspace{0.1cm}
    \caption{Projected Compression (PC) vs Hard Pruning Retraining (HPR) - Cross-Entropy loss of base model compression of different sizes (Model $N$) pretrained with 80:1 tokens to parameters ratio. Best loss from each method pair is bolded. }
    \label{tab:80tpp}
\end{table}

\subsection{Base model size and quality}

\begin{figure}
    \centering
    \hfill
    \centering
    \begin{minipage}[t]{0.48\textwidth}
        \includegraphics[width=\linewidth]{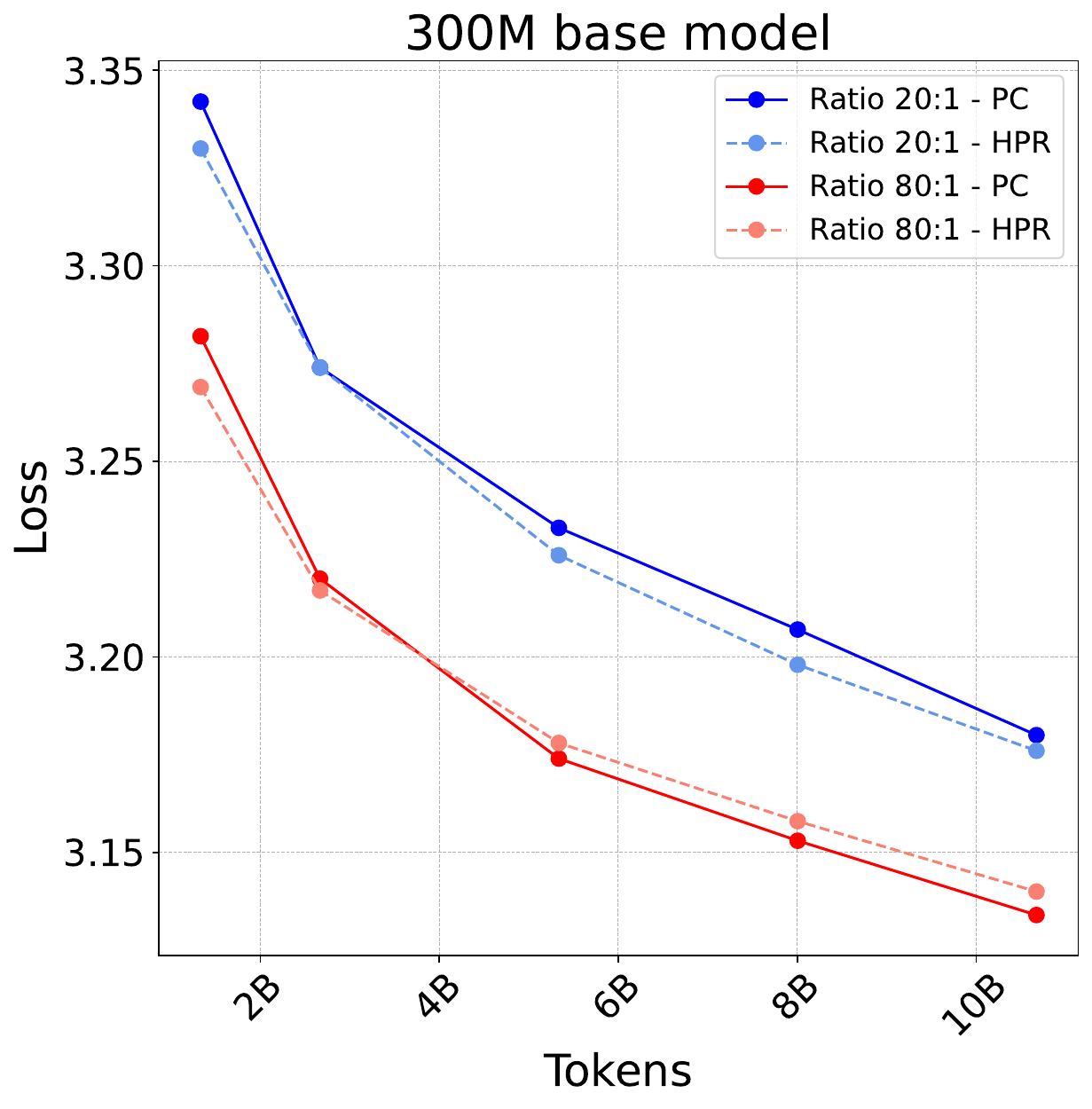}
        \label{fig:300M}
    \end{minipage}
    \hfill
    \begin{minipage}[t]{0.48\textwidth}
        \includegraphics[width=\linewidth]{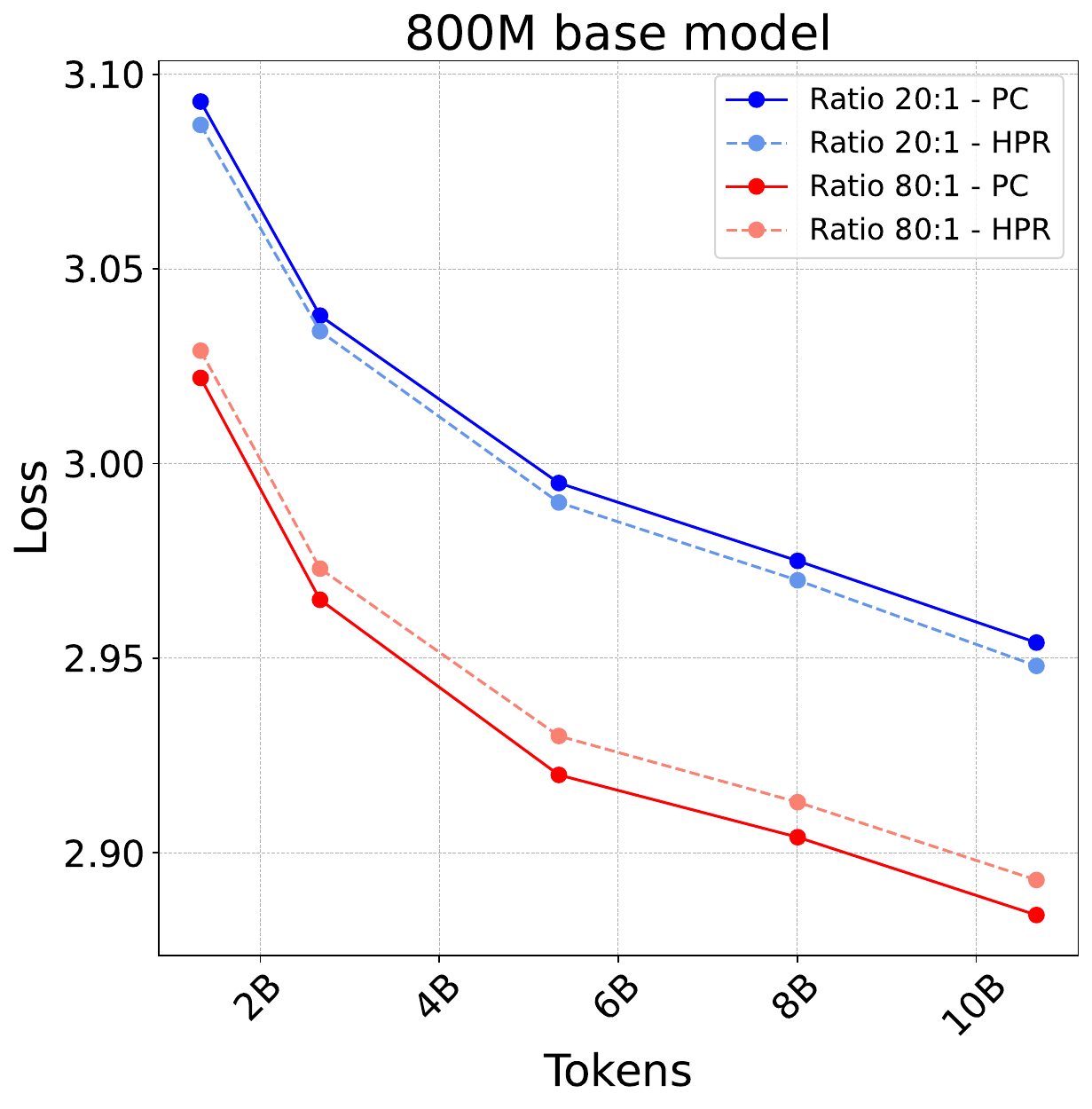}
        \label{fig:800M}
    \end{minipage}
   \caption{Projected Compression (PC) vs Hard Pruning Retraining (HPR) for base models of different sizes pretrained with different tokens to parameters ratio (Ratio $T$:$N$).}
    \label{fig:pc_vs_hp}
\end{figure}

The experimental trends suggest that Projected Compression is particularly well-suited for compressing larger and higher-quality base models. In particular, PC performs best when the model has been trained on a sufficiently large number of tokens relative to its parameter count. This is illustrated in \Cref{fig:pc_vs_hp}, where Projected Compression yields clear improvements over hard pruning at a token-to-parameter ratio of 80:1. Similarly, the performance advantage of the PC is more pronounced in the 800M-parameter model than in the 300M-parameter variant, as shown in \Cref{tab:80tpp}. 

These observations support the view that Projected Compression is a more effective compression method for the types of models that are the most efficient and valuable to compress. This efficiency trend that compression is most beneficial when applied to large, high-quality base models in shown in Frantar et al. work \citep{frantar2023scalinglawssparselyconnectedfoundation}.

\section{Conclusion and Future Work}
In this work, we presented Projected Compression, a novel compression technique that preserves access to all original model parameters through learnable projection weights. Experimental results demonstrate that our method is an effective method for compressing standard transformer-based models.
This work opens the possibility of adapting PC method to a broader range of architectures, which we leave for future exploration.

\clearpage
\section{References}
\bibliographystyle{unsrt}

\bibliography{bib} 

\newpage
\appendix

\section{Experiments}

\subsection{Weights importances in experiments initialization}

We have tested the influence on the importance of random and magnitude-based weights for tested base models in both tested compression methods. The magnitude-based method produced the worst performance than random pruning for each computation method \Cref{tab:mag}.

\subsection{All experiments}
\label{appx:a}

\begin{table}[ht]
    \centering
\begin{tabular}{ccc}
\toprule 
Model size ($N$) & Tokens ratio ($T:N$) & Loss \\
\midrule
800M&80:1&2.7234\\
800M&20:1&2.8547\\
300M&80:1&3.0041\\
300M&20:1&3.1450\\

\end{tabular}
\vspace{0.1cm}
    \caption{Base models loss. }
    \label{tab:base}
\end{table}

\begin{table}[ht]
    \centering
\begin{tabular}{ccccc}
\toprule 
 & \multicolumn{2}{c}{Projected Compression} & \multicolumn{2}{c}{Hard Pruning Retraining} \\
\midrule
Retraining tokens & \texttt{2.5B} & \texttt{5B} & \texttt{2.5B} & \texttt{5B}\\
\midrule
\text{random} & $\textbf{2.9652}$ & $\textbf{2.9203}$ & $\textbf{2.9733}$ & $\textbf{2.9300}$ \\
\midrule
\text{magnitude} & $2.9655$ &  $2.9213$ & $2.9764$ &  $2.9320$  \\
\midrule
\end{tabular}
\vspace{0.1cm}
    \caption{Magnitude vs random based weights importance. Cross-Entropy loss when compressing 800M model by 50\% (80:1 tokens to parameters ratio). }
    \label{tab:mag}
\end{table}

\begin{table}[ht]
    \centering
\begin{tabular}{ccccccccc}
\toprule  
& & &  \multicolumn{5}{c}{Tokens processed during training compression pipeline}\\
Model $N$ & Ratio $T$:$N$ & Method &  \texttt{1.25B} & \texttt{2.5B} & \texttt{5B}& \texttt{7.5B} & \texttt{10B}\\
\bottomrule
\multicolumn{8}{c}{35\% compression}\\
\midrule
\text{300M} & \text{20:1} & \text{HPR} & \textbf{3.2645} & \textbf{3.2175} & \textbf{3.1816} & \textbf{3.1570} & \textbf{3.1339}\\
\text{300M} & \text{20:1} & \text{PC} & 3.2683 & 3.2221 & 3.1845 & 3.1622 & 3.1400\\
\hdashline
\text{300M} & \text{80:1} & \text{HPR} & $\textbf{3.1759}$ & $3.1362$ & $3.1070$ & $3.0925$ & $3.0775$\\
\text{300M} & \text{80:1} & \text{PC} & $3.1776$ & $\textbf{3.1361}$ & $\textbf{3.1037}$ & $\textbf{3.0882}$ & $\textbf{3.0711}$\\
\hdashline
\text{800M} & \text{20:1} & \text{HPR} & \textbf{3.0074} & \textbf{2.9657} & \textbf{2.931} & \textbf{2.9179} & \textbf{2.9031}\\
\text{800M} & \text{20:1} & \text{PC} & 3.0152 & 2.974 & 2.9380 & 2.9238 & 2.9056\\
\hdashline
\text{800M} & \text{80:1} & \text{HPR} & $\textbf{2.9224}$ & $2.8827$ & $2.8519$ & $2.8420$ & $2.8259$\\
\text{800M} & \text{80:1} & \text{PC} & $3.0327$ & $\textbf{2.8780}$ & $\textbf{2.8465}$ & $\textbf{2.8355}$ & $\textbf{2.8192}$\\

\toprule 
\multicolumn{8}{c}{50\% compression}\\
\midrule
\text{300M} & \text{20:1} & \text{HPR} & \textbf{3.3302} & \textbf{3.2741} & \textbf{3.2258} & \textbf{3.1980} & \textbf{3.1758}\\
\text{300M} & \text{20:1} & \text{PC} & 3.3421 & 3.2742 & 3.2329 & 3.2070 & 3.1830\\
\hdashline
\text{300M} & \text{80:1} & \text{HPR} & $\textbf{3.2688}$ & $\textbf{3.2172}$ & $3.1781$ & $3.1577$ & $3.1402$\\
\text{300M} & \text{80:1} & \text{PC} & $3.2823$ & $3.2198$ & $\textbf{3.1744}$ & $\textbf{3.1528}$ & $\textbf{3.1341}$\\
\hdashline
\text{800M} & \text{20:1} & \text{HPR} & \textbf{3.0872} & \textbf{3.0345} & \textbf{2.9896} & \textbf{2.9695} & \textbf{2.9484}\\
\text{800M} & \text{20:1} & \text{PC} & 3.0925 & 3.0380 & 2.9948 & 2.9752 & 2.9543\\
\hdashline
\text{800M} & \text{80:1} & \text{HPR} & $3.0286$ & $2.9733$ & $2.9300$ & $2.9128$ & $2.8930$\\
\text{800M} & \text{80:1} & \text{PC} & $\textbf{3,0217}$ & $\textbf{2.9652}$ & $\textbf{2.9203}$ & $\textbf{2.9037}$ & $\textbf{2.8840}$\\

\toprule 
\multicolumn{8}{c}{65\% compression}\\
\midrule
\text{300M} & \text{20:1} & \text{HPR} & \textbf{3.4125} & \textbf{3.3464} & \textbf{3.2919} & \textbf{3.2628} & \textbf{3.2455}\\
\text{300M} & \text{20:1} & \text{PC} & 3.4241 & 3.356 & 3.3027 & 3.2755 & 3.2523\\
\hdashline
\text{300M} & \text{80:1} & \text{HPR} & $\textbf{3.3753}$ & $\textbf{3.3095}$ & $3.2603$ & $3.2392$ & $3.2175$\\
\text{300M} & \text{80:1} & \text{PC} & $3.3858$ & $3.3134$ & $\textbf{3.2591}$ & $\textbf{3.2341}$ & $\textbf{3.2133}$\\
\hdashline
\text{800M} & \text{20:1} & \text{HPR} & \textbf{3.1733} & \textbf{3.1098} & \textbf{3.0577} & \textbf{3.0341} & \textbf{3.0109}\\
\text{800M} & \text{20:1} & \text{PC} & 3.1848 & 3.1188 & 3.0638 & 3.0380 & 3.0159\\
\hdashline
\text{800M} & \text{80:1} & \text{HPR} & $\textbf{3.1245}$ & $\textbf{3.0569}$ & $\textbf{3.0056}$ & $2.9828$ & $2.9622$\\
\text{800M} & \text{80:1} & \text{PC} & $3.1375$ & $3.0626$ & $3.0069$ & $\textbf{2.9815}$ & $\textbf{2.9591}$\\
\bottomrule
\end{tabular}
\vspace{0.1cm}
    \caption{Projected Compression (PC) vs Hard Pruning Retraining (HPR) - Cross-Entropy loss of base model compression of different sizes (Model $N$) pretrained with different tokens to parameters ratio (Ratio $T$:$N$). Best loss from each method pair is bolded. }
    \label{tab:hp_mag}
\end{table}

\end{document}